# MIST: Missing Person Intelligence Synthesis Toolkit

Elham Shaabani, Hamidreza Alvari, and Paulo Shakarian*
Arizona State University
Tempe, AZ 85281
{shaabani, halvari, shak}@asu.edu

J.E. Kelly Snyder
Find Me Group
Chandler, AZ 85249
kelly@findmegroup.org

## ABSTRACT

Each day, approximately 500 missing persons cases occur that go unsolved/unresolved in the United States. The non-profit organization known as the Find Me Group (FMG), led by former law enforcement professionals, is dedicated to solving or resolving these cases. This paper introduces the Missing Person Intelligence Synthesis Toolkit (MIST) which leverages a data-driven variant of geospatial abductive inference. This system takes search locations provided by a group of experts and rank-orders them based on the probability assigned to areas based on the prior performance of the experts taken as a group. We evaluate our approach compared to the current practices employed by the Find Me Group and found it significantly reduces the search area - leading to a reduction of 31 square miles over 24 cases we examined in our experiments. Currently, we are using MIST to aid the Find Me Group in an active missing person case.

## 1. INTRODUCTION

Each day, approximately 500 missing persons cases occur that go unsolved/unresolved in the United States. The non-profit organization known as the Find Me Group (FMG), led by former law enforcement professionals, is dedicated to solving or resolving these cases. This non-profit operates with limited resources - so it must use its volunteer assets in a highly efficient manner. This paper introduces the Missing Person Intelligence Synthesis Toolkit (MIST) which leverages a data-driven variant of geospatial abductive inference [24]. This system takes search locations provided by a group of experts and rank-orders them based on the probability assigned to areas based on the prior performance of the experts taken as a group. We evaluate our approach compared to the current practices employed by the FMG and found it significantly reduces the search area. In 24 cases examined in our experiments (on real-world data provided by FMG), we found our approach to be able to reduce total search area by a total of 31 square miles for standard searches and by 19 square miles when dog team assets obtain a detection. This reduction is significant for the following reasons:

- **Reduction in time to locate missing persons.** In cases where baseline provided 20 square miles or more (the most difficult cases), we achieved reduction in search area of 7 to 56 square miles. As 3-5 square miles are searched on a typical day (terrain dependent), such a reduction can potentially increase the chance of a missing person being found alive.

- **Reduction in direct costs.** During a search, FMG spends approximately $2200 per day. In all tests, our approach reduced the search area in the majority of cases which can be interpreted as a reduction in direct costs.

- **Reduction in indirect costs.** FMG relies extensively on volunteers to augment searches. During searches, these individuals often lose earnings from their day job or small business. As many volunteers also perform consulting or other services to law enforcement, longer searches lead to loss of revenue and opportunity. In one case, a volunteer estimated a loss of $15K. Again, our approach leads to a consistent reduction in search area - hence reducing these indirect costs.

Specifically, we contribute an extension to geospatial abduction [24] that leverages historical data of individual experts. We also create new algorithms to learn parameters of a geospatial abduction model from data based on integer programming. We then evaluate these algorithms on real-world data provided by the FMG under a variety of different settings. This approach learns pattern of each reporter independently and is able to overcome outliers if any. It also does well on the limited data. This work has prepared us in our ongoing deployment of the software. At the time of this writing, we have provided results of MIST to support an active case with FMG. Figure 1 shows an example output of MIST where it rank-orders search locations. FMG is currently using this information to support their operations. They found the result consistent with their experiences.

---
*U.S. Provisional Patent 62/345,193. Contact shak@asu.edu for licensing information.

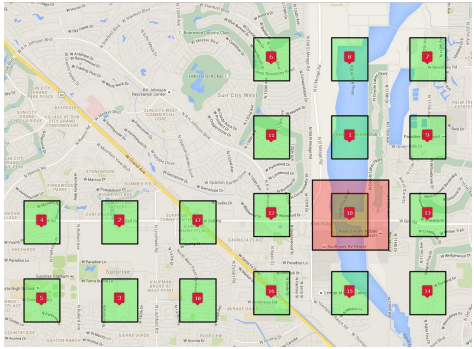

Figure 1: Mapping of ordered grids by MIST (green squares) and current searched area by FMG (red square).

The rest of the paper is organized as follows. In Section 2, we present the background of the missing person problem. Next, we provide the technical preliminaries. We discuss our data-driven extension in Section 4. In Section 5, we detail our algorithmic approach. We introduce our dataset and conduct data analysis in Section 6. Next, we discuss the experimental results in Section 7. We review the related work in Section 8. We conclude the paper presenting future research directions.

## 2. BACKGROUND

Missing persons cases have been on the rise in the USA for the past twenty years. Currently, approximately 4000 people go missing each and every day. Approximately 3500 of those cases are solved or resolved (i.e., cases solved by only providing accurate information to the authorities and without physical involvement), which leaves an astounding number of victims that are never located. In the case of missing adults 13 years of age and older, the police are not required or obligated to conduct an investigation or search unless there are extenuating circumstances such as suicide, a potential for violence, medical reasons, etc. This leaves families and friends without professional assistance in locating their loved ones. The Find Me Group (FMG) was founded by retired U.S. Drug Enforcement Agency (DEA) Special Agent J.E. "Kelly" Snyder in 2002. The group consists of current and retired law enforcement officers with a wide-range of investigative expertise, including but not limited to linguistics, handwriting analysis, body language, missing person/homicide experience and search-and-rescue field management skills. The FMG has trained experts/sources that provide detailed location information where missing individuals can be found. Many of these experts have the ability to provide GPS coordinates to locate missing persons with a varying levels of success. The FMG focus/goal is to provide accurate location information in a timely manner and minimize the potential of finding the victim deceased. Thirty canine handlers certified in tracking, scent and cadaver complements the FMG and has led to many instances where the person in questions was located.

Equally disturbing nationwide is the rise in human trafficking, which aligns within the missing person category. This type of crime has long-term and devastating results. The work of this paper is also the first step toward an all-encompassing methodology of identifying locations of missing persons who were victims of human trafficking. Another important related crime is homicide. Many missing persons and human trafficking victims are found deceased due to this crime. This work represents initial progress in aiding toward crimes of this nature as well.

In this paper, we formulate the problem of "finding missing person" with respect to information provided by FMG's experts, formally as a variant of the geospatial abduction problem (GAP) [22]. To account for the key nuances of "finding missing person" problem though, we extended the GAP framework to better suite this domain. In particular, we extend the GAP formalism with a data driven model - accounting for the previous performance of experts aiding in the missing person cases. We list the unique characteristics of our framework here. Later in the next section, we provide our technical approach to each.

1. **Explanation Size.** One key difference "finding missing person" problem has from other GAP instances, is that the explanation (the result of a GAP inference algorithm) only consists of a single related location (i.e., the location of the missing person) corresponding to the phenomenon under study. This differs from returning a set of $k$ locations in the previously-introduced GAP formalisms. Consequently, here, an explanation will consist of a single point, which in turn led us to explore a non-deterministic version of the original explanation.

2. **Distance Constraints.** In the original GAP formalism, each observed geospatial phenomenon is related to unobserved "partner" points through a distance constraint - $(\alpha,\beta)$ where $\alpha$ is the minimum distance between an observation and partner and $\beta$ is the maximum distance. As described, this pair of constraints was the same for *all* observations. However, in the missing persons problem, each observation corresponds to a different domain expert - and hence has a different $(\alpha,\beta)$ constraint pair. Further, we study how this is best learned from data, as well as "soften" the constraint - assigning a probability of the partner point being less than $\alpha$, between distances $\alpha$ and $\beta$, and greater than distance $\beta$ from an observation.

3. **Uncertainty.** As we learn the $(\alpha,\beta)$ distance constraints for each observation and associate corresponding probabilities from historical data, it makes sense that the inference step is treated probabilistically - which differs from the original deterministic GAP framework. Further, this enables us to rank the potential partner locations (again, as an explanation consists of one point, ranking search locations is more useful in a practical sense).

4. **Independent Observations.** In the original GAP framework, independence amongst the observations was not an assumption in the framework. However, FMG compartmentalizes the information from their law enforcement experts from one another in a manner to obtain independent reporting. Hence, we make this assumption in this paper and it is supported by our experimental results.

FMG currently uses a simple heuristic to rank-order potential search locations for a missing person (we describe this later in Section 5). Once ranked, FMG leverages a variety of assets. Figure 2 depicts a recently searched area for a case. It represents a screen shot of the tracks from the GPS units that the dogs wear as well as the handheld units that the searchers wear. This shows several dog tracks and the human tracks. The green, dark blue, magenta represent three dogs, the grey and red represent two human searchers. The teal track is a trailing dog, ascertaining a direction of travel. The straight lines tend to be humans and the rapidly changing direction lines are dogs as they grid around the humans. Figure 3 shows real-world examples of how the FMG practices in an undisclosed location.

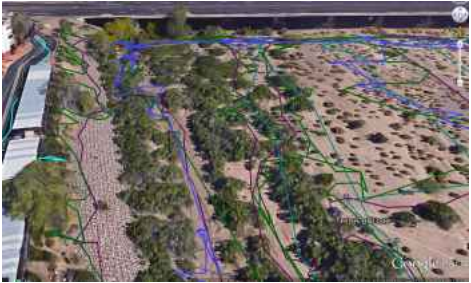

Figure 2: Screen shot of the tracks from the GPS units.

## 3. TECHNICAL PRELIMINARIES

In this section, we briefly explain geospatial abductive inference [24], and introduce our new (introduced in this paper) data-driven probabilistic extension. We show how this extension was used to address the unique characteristics of the missing person location problem.

In general, *abduction* or *abductive inference* [12] refers to a type of logic or reasoning to derive plausible explanations for a given set of facts [13]. Abduction has been extensively studied in medicine [13, 14], fault diagnosis [3], belief revision [11], database updates [8, 4] and AI planning [5]. Two major existing theories of abduction include logic-based abduction [6] and set-covering abduction [2]. Though none of the above papers takes into account spatial inference, [25] presents a logical formalism dealing with objects' spatial occupancy, while [18] describes the construction of a qualitative spatial reasoning system based on sensor data from a mobile robot.

*Geospatial abduction problem* (GAP) [22], on the other hand, refers to the problem of identifying unobserved partner locations (i.e., the location of a missing person) that best explain a set of observed phenomenon with known geographic locations. *Geospatial abduction* was first introduced in [23] and later extended in [24, 21, 20, 19]. More formally, each GAP consists of three major elements [22]: (1) observations: a set of observations that explain the locations associated with the event under study (e.g., in this application, the locations reported by the domain experts), (2) distance constraints: a pair $(\alpha, \beta) \in \mathbb{R}$ corresponding to lower and upper bounds on the distances between observation and partner location and, (3) feasibility predicate: this allows to specify whether an area on the map is a potential location for a partner.

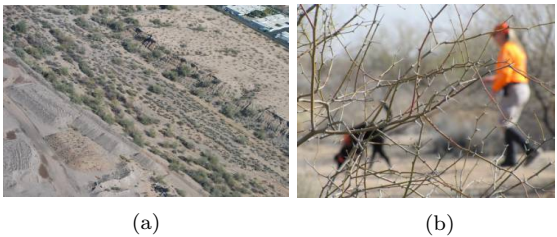

Figure 3: (a) Picture of the search area taken from the plane. (b) Search team.

Next, we present the notations and definitions used throughout the paper, and review the geospatial abduction framework of [22]. In the next section, we describe specialized extensions that were necessary to study our problem. First, without loss of generality, we assume throughout the paper that a map (resp. space) is represented by a discrete two dimensional grid of size $M \times N$, defined as follows:

DEFINITION 3.1. **(Space).** *Given natural numbers $M$, $N$, the space $\mathcal{S}$ is the set $[1, \ldots, M] \times [1, \ldots, N]$.*

Associated with the space is a *distance function* $d : \mathcal{S} \times \mathcal{S} \to \mathbb{R}^+$ that satisfies the normal distance axioms: $d(p_i, p_i) = 0$, $d(p_i, p_j) = d(p_j, p_i)$, and $d(p_i, p_j) \leq d(p_i, p_q) + d(p_q, p_j)$.

Note that we use $o$ to represent the observer (source of information) and $p_o$ to represent the location he/she reported (which differs slightly from the original framework). From these observations (reports), the corresponding unobserved phenomenon is the actual location of the missing person. In the original framework, the explanation consisted of geographic locations that were located at least distance $\alpha$ and no more than distance $\beta$ away from each observation. In this work, we generalize this notion by providing $\alpha, \beta$ pair for each observer - denoted $\alpha_o, \beta_o$.

DEFINITION 3.2. **(Feasibility Function).** *A feasibility function feas is defined as feas : $\mathcal{S} \to \{\text{True}, \text{False}\}$.*

A key use for the feasibility function here is for an initial reduction of the search space by the FMG. This is due to the fact that missing person reports often span a large area and an initial reduction is necessary for practical reasons. An obvious future direction would be to utilize a probabilistic variant of the feasibility function - which would assign a prior probability to a location for a missing person. However, in this application, it is unclear where such a distribution would come from. Further, as the search space is relatively large when compared to FMG resources, the deterministic version of this definition is more appropriate for operational reasons.

Due to resource constraints and the generally large areas over which reports are spread, FMG typically only searches areas for which there is a report. As we shall describe in Section 5, they search a $1 \times 1$ mile square surrounding a location reported by an observer. As such is the case, we shall assume the following feasibility function throughout this paper:

$$\text{feas}(p) = \begin{cases} \text{True} & \text{if } p \in \mathcal{O} \\ \text{False} & \text{otherwise} \end{cases} \quad (1)$$

Unless otherwise noted, we shall assume the above function is used for feasibility and hence the subset of the space considered will be the points in $\mathcal{O}$.

We now come to the important definition of an *explanation*. Intuitively, for a given set of points $\{p_1, \ldots, p_{|\mathcal{O}|}\}$

reported by observers in $\mathcal{O}$, an explanation is a set of points $\mathcal{E}$ such that every point in this set is feasible and for every observation, there is a point in $\mathcal{E}$ that is at least $\alpha$ units away from the observation, but no more than $\beta$ units from the observation.

DEFINITION 3.3. (($\alpha,\beta$) **Explanation**). *Suppose $\mathcal{O}$ is the set of observations, $\mathcal{E}$ is a finite set of points in $\mathcal{S}$, and $0 \leq \alpha, \beta \leq 1$ are two real numbers. $\mathcal{E}$ is said to be an $(\alpha, \beta)$ explanation of $\mathcal{O}$ iff:*

- *$p \in \mathcal{E}$ implies that feas$(p) =$ True, i.e., all points in $\mathcal{E}$ are feasible.*
- *$(\forall o \in \mathcal{O})(\exists p \in \mathcal{E})\ \alpha \leq d(p,o) \leq \beta$, i.e., every observation is neither too close nor too far from some point in $\mathcal{E}$.*

Thus, an $(\alpha,\beta)$ explanation is a set of points. Each point must be feasible and every observation must have an analogous point in the explanation which is neither too close nor too far.

Again, we note that here an explanation will consist of a single point - the location of the missing person. Hence, this deterministic definition of an explanation will not suffice - as in practice there will often not exist an explanation for a given problem instance. As such is the case, we extended this framework using a data-driven approach.

## 4. DATA-DRIVEN EXTENSIONS

In this section, we describe our data-driven probabilistic extension to the original GAP formalism. The framework extensions in this section were not previously introduced and are new in this paper. In order to do so, we first introduce some preliminary notation. For point $p \in \mathcal{S}$, the random variable $\mathcal{P}_p$ denotes that the missing person was found at point $p$, so this is either true or false. We will use $\mathcal{P}_p$ as shorthand for $\mathcal{P}_p =$ True. For observer $o \in \mathcal{O}$ the random variable $\mathsf{O}_o$ can be assigned to one of the points in $p$. Based on this notation, we define an *explanation distribution*.

DEFINITION 4.1 (**Explanation Distribution**). *Given a set of observers $\mathcal{O}$ and a set of reported locations by each observer $p_1, \ldots, p_o, \ldots, p_{|\mathcal{O}|}$, an **explanation distribution** is a probability distribution over all points in $\mathcal{S}$ - directly addressing characteristic 3 of this application (see Section 2). This distribution assigns the probability of a missing person being located at each point conditioned on the observers reporting their respective locations. Formally, it is written as $Pr(\mathcal{P}_p|\bigwedge_{o \in \mathcal{O}} \mathsf{O}_o = p_o)$.*

The key intuition is that if we are able to compute an explanation distribution, we can then rank-order points in the space by probability - and hence conserve search resources. Note that the explanation distribution is over all points - implying that there is precisely one location. While generalizations that allow for more than one location are possible in such a probabilistic framework, we keep the size at one due to the first characteristic of our problem (as described in Section 2).

In this paper, we make an assumption of *distance primacy* meaning the distance constraints $(\alpha_o, \beta_o)$ relate the $\mathcal{P}_p$ with $\bigwedge_{o \in \mathcal{O}} \mathsf{O}_o = p_o$. Hence, we introduce another random variable, $\mathfrak{R}^{\beta_o}_{p,p'}$ which is true if $d(p,p') \leq \beta_o$ and false otherwise.

Note that in the remainder of this section, we will use one distance constraint ($\beta$) for sake of brevity - though this idea can be extended for multiple distance constraints (as per characteristic 2 from Section 2). In fact, we leverage multiple distance constraints in our optimization procedure for parameter selection introduced later. Hence, by distance primacy, we have the following relationships.

$$Pr(\mathcal{P}_p | \bigwedge_{o \in \mathcal{O}} \mathsf{O}_o = p_o) = Pr(\mathcal{P}_p | \bigwedge_{o \in \mathcal{O}} \mathfrak{R}^{\beta}_{p,p_o}) \quad (2)$$

By Bayes' Theorem, this is equivalent to the following.

$$\frac{Pr(\mathcal{P}_p) \times Pr(\bigwedge_{o \in \mathcal{O}} \mathfrak{R}^{\beta}_{p,p_o} | \mathcal{P}_p)}{Pr(\bigwedge_{o \in \mathcal{O}} \mathfrak{R}^{\beta}_{p,p_o})} \quad (3)$$

However, by characteristic 4, we assume that the observers report information independently, which gives us the following.

$$\frac{Pr(\mathcal{P}_p) \times \prod_{o \in \mathcal{O}} Pr(\mathfrak{R}^{\beta}_{p,p_o} | \mathcal{P}_p)}{Pr(\bigwedge_{o \in \mathcal{O}} \mathfrak{R}^{\beta}_{p,p_o})} \quad (4)$$

Due to our application, we will not consider the prior probability $Pr(\mathcal{P}_p)$ as each missing person case occurs in a different geographic location - and due to the wide range of cases that span multiple countries, data supporting a realistic, informed prior is highly sparse. As such, we consider a uninformed prior. Further, for notational simplicity, we shall use the notation $\rho^{\beta}_o$ for the quantity $Pr(\mathfrak{R}^{\beta}_{p,p_o} = $ True$|\mathcal{P}_p = $ True$)$. Therefore, we can rank points in the space based on the explanation distribution by simply considering their log-likelihood computed as follows:

$$\sum_{\substack{o \in \mathcal{O} \\ d(p,p_o) \leq \beta}} \log(\rho^{\beta}_o) + \sum_{\substack{o \in \mathcal{O} \\ d(p,p_o) > \beta}} \log(1 - \rho^{\beta}_o) \quad (5)$$

Hence, the inference step for this problem is straightforward provided we know the values $\beta$ and $\rho^{\beta}_o$ for each observer $o \in \mathcal{O}$ (or similar parameters if considering more than one distance constraint). If we know the value $\beta$ we can then compute $\rho^{\beta}_o$ based on a corpus of historical data concerning the accuracy of reporter $o$. Given a corpus of previous cases for the observer $C_o$ where the found location was $p^c$ and the location reported by the observer was $p^c_o$, we can compute $\rho^{\beta}_o$ as follows:

$$\rho^{\beta}_o = \frac{|\{c \in C_o\ s.t.\ d(p^c, p^c_o) \leq \beta\}|}{|C_o|} \quad (6)$$

Hence, we also adjust $\rho^{\beta}_o$ to account for volume of the reporter's history to provide the effect of regularization. Considering $\eta_o$ as the portion of total number of cases in which observer $o$ has participated, to the total number of cases, and $\epsilon$ as a non-negative parameter, we define $\rho^{\beta,\epsilon}_o$ as follows:

$$\rho^{\beta,\epsilon}_o = \rho^{\beta}_o - \epsilon \times (1 - \eta_o) \quad (7)$$

The situation is further complicated with multiple distance constraints. We propose an optimization approach to this problem in the next section.

## 5. ALGORITHMIC APPROACH

In this section, we present our algorithmic approach to special case of geospatial abductive inference. First, we ex-plain the method that FMG currently uses. Then, we provide our proposed optimization approach to solve the problem.

## 5.1 Existing Method

The FMG uses the following method to explore the missing person location. Given the reported locations provided by different observers, FMG initially creates a search area (grid) as follows. First, they draw building blocks (or boxes) of size 1×1 mile centered at each reported location (note that depending on the situation, these boxes may overlap). Then, they search the entire grid in the following order. First, they search the larger areas created of the overlapping boxes, and if the missing person was not found, they explore the remaining boxes in the order of the observers' history (how well they did in the past). The whole process is repeated by extending the size of boxes to 2×2 miles, if the missing person was not located. Note that we use the same grid in our proposed methods.

## 5.2 Proposed Methods

As described, for simplicity, we first elaborate on the required steps to calculate the best $\beta_o$ for each observer. Then, we extend the idea for multiple distance constraints. Let $[\beta_o]$ be the set of possible error radii. Note that for $C_o$ cases where observer reported a location, there are at most $|C_o|$ possible values for $\beta_o$. Hence, our goal is to select as a set of these distance constraints - one for each observer. We do this through an integer program - where for each observer $o \in \mathcal{O}$ and each associated distance constraint $\beta_o \in [\beta_o]$ we have an indicator variable $X_{o,\beta_o}$ that is 1 if we use that value and zero otherwise. We shall refer to this as the *single constraint integer program*. Hence, we find an assignment of values to these indicator variables in order to maximize the following quantity:

$$F_1 = \sum_{c \in C} \sum_{o \in \mathcal{O}} \sum_{\beta \in [\beta_o]} \Big[ \delta_\beta(p^c, p_o^c) \times \log \rho_o^\beta \times X_{o,\beta} +$$
$$(1 - \delta_\beta(p^c, p_o^c)) \times \log(1 - \rho_o^\beta) \times X_{o,\beta} \Big] \quad (8)$$

subject to the following constraints:

$$\forall X_{o,\beta} \in \{0, 1\} \quad (9)$$

$$\forall o, \sum_{\beta \in [\beta_o]} X_{o,\beta} \leq 1 \quad (10)$$

$$\sum_o \sum_{\beta \in [\beta_o]} X_{o,\beta} = k \quad (11)$$

where $k$ is a cardinality that limits the number of reporters (which is set to a natural number in the range $1, \ldots, |\mathcal{O}|$), and $\delta_\beta(x, y)$ is defined as:

$$\delta_\beta(x, y) = \begin{cases} 1 & \text{if } d(x, y) \leq \beta \\ 0 & \text{if } d(x, y) > \beta \end{cases} \quad (12)$$

However, this equation will result in tendency toward selecting the largest distance constraints. This has the effect of not only maximizing the probability of the locations where the missing person was found, but also can increase the probability of other locations. Intuitively, we want to also minimize the following quantity:

$$F_2 = \sum_{c \in C} \sum_{o \in \mathcal{O}} \sum_{p \in \{\mathcal{S} \setminus p^c\}} \sum_{\beta \in [\beta_o]} \Big[ \delta_\beta(p, p_o^c) \times \log \rho_o^{'\beta} \times X_{o,\beta}$$
$$+ (1 - \delta_\beta(p, p_o^c)) \times \log(1 - \rho_o^{'\beta}) \times X_{o,\beta} \Big] \quad (13)$$

Therefore, the objective function we seek to optimize is

$$L_1 = \max(F_1 - F_2) \quad (14)$$

THEOREM 5.1. *Number of variables in single-distance constraint integer program is $O(\text{avg}(|C_o|) \cdot |\mathcal{O}|)$.*

We extend the previous formulation by allowing the objective function to find a pair of distance constraints for each reporter. We have experimentally found diminishing returns on performance (and in many cases increased complexity) with more than two constraints. This will give us the *double distance constraint integer program* as follows:

$$F_1' = \sum_{c \in C} \sum_{o \in \mathcal{O}} \sum_{\alpha \in [\beta_o]} \sum_{\substack{\beta \in [\beta_o] \\ \beta \geq \alpha}} \Big[ \delta_\alpha(p^c, p_o^c) \times \log \rho_o^{\alpha,\epsilon} \times X_{o,\alpha,\beta}$$
$$+ \Big(1 - \delta_\alpha(p^c, p_o^c)\Big) \times \delta_\beta(p^c, p_o^c) \times \log\Big(\rho_o^{\beta,\epsilon} - \rho_o^{\alpha,\epsilon}\Big) \times X_{o,\alpha,\beta} +$$
$$(1 - \delta_\beta(p^c, p_o^c)) \times \log\Big(1 - \rho_o^{\beta,\epsilon}\Big) \times X_{o,\alpha,\beta} \Big]$$

subject to the following constraints:

$$\forall X_{o,\alpha,\beta} \in \{0, 1\}$$

$$\forall o, \sum_{\alpha,\beta \in [\beta_o]} X_{o,\alpha,\beta} \leq 1$$

Likewise, we use the following objective function, to avoid bias toward selecting the largest $\beta$'s.

$$L_2 = \max(F_1' - F_2') \quad (15)$$

where $F_2'$ is defined as follows:

$$F_2' = \sum_{c \in C} \sum_{o \in \mathcal{O}} \sum_{\alpha \in [\beta_o]} \sum_{\substack{\beta \in [\beta_o] \\ \beta \geq \alpha}} \Big[ \sum_{p \in \{\mathcal{S} \setminus p^c\}} \delta_\alpha(p, p_o^c) \times \log \rho_o^{'\alpha,\epsilon} \times X_{o,\alpha,\beta} +$$
$$\Big(1 - \delta_\alpha(p, p_o^c)\Big) \times \delta_\beta(p, p_o^c) \times$$
$$\log(\rho_o^{'\beta,\epsilon} - \rho_o^{'\alpha,\epsilon}) \times X_{o,\alpha,\beta} +$$
$$\Big(1 - \delta_\beta(p, p_o^c)\Big) \times \log\Big(1 - \rho_o^{'\beta,\epsilon}\Big) \times X_{o,\alpha,\beta} \Big] \quad (16)$$

THEOREM 5.2. *Number of variables in double distance constraint integer program is $O(\text{avg}(|C_o|)^2 \cdot |\mathcal{O}|)$.*

While we obtained a significant reduction in the area searched by setting the cardinality constraint $k = \mathcal{O}$, we found that varying it would often lead to further improvement. We gradually increased the number of observers from one to the total number of observers and each time, we learned the distance constraints for the last added observers. In this method of optimization, we may choose a specific number of points in each iteration. The number of points added with each iteration can be determined based on available resources.

We also defined two heuristic to discriminate points with the same probability. In each iteration, we chose the point with highest probability. If there were more than one point, we applied following heuristics: (1) we chose the points which had most of the reported locations in its $1 \times 1$ mile. (2) we chose the point which had the maximum summation of the priors of the reporters in its $1 \times 1$ mile.

Algorithm 1 is a specific variant of restricted model. In this algorithm, in each iteration one point (i.e., representative of a $1 \times 1$ mile) is selected. Though we note that this can easily be adjusted in practice. If the area size we are able to search is larger than number of observers, we sort the representatives based on their probabilities. Then, we apply two heuristics to rank them (similar to Lines 11-19 ).

---

**Algorithm 1** Iterative Search Resource Allocation

1: **procedure** OPT-POINT-BY-POINT($A, c, \mathcal{S}, \rho$)  ▷ Train set $A$, Test case $c$
2:    List $R = \emptyset$                ▷ Output
3:    **for** $k \in [1, |\mathcal{O}_c|]$ **do**  ▷ $k$ is a constant value of the constraint
4:       Find assignment of variables that optimize (15) w.r.t. (9 - 11)
5:       $RP \leftarrow$ Order by (5)        ▷ Ranked points RP
6:       $RP \leftarrow RP \setminus R$
7:       Pick $P \subseteq RP$ with largest probabilities
8:       **if** $P$ includes one point **then**
9:          $R = R \cup P$
10:      **else**
11:         $p \leftarrow Heuristic(P)$
12:         $R = R \cup \{p\}$
13:    **return** $R$

---

THEOREM 5.3. *The time complexity of the algorithm (1) is $O(|C| \cdot \text{avg}(|C_o|)^2 \cdot \text{avg}(|\mathcal{O}_c|)^3)$.*

## 6. MISSING PERSON DATASET

In this section, we describe our dataset and briefly discuss the observation made from our initial data analysis.

### 6.1 Overview

Our dataset includes cases (i.e., missing persons), found status (alive/deceased), found location (latitude and longitude), age and reason for disappearance as well as the potential locations (latitude and longitude) associated with the reporters/experts. The description of this dataset is summarized in Table 1. Note that in some cases, we are aware of reports, but do not have the found location ($p_o^c$). In this work, we only have 29 cases with the known found locations used for the experiments. However, for the data analysis, the entire dataset is applied.

### 6.2 Data Analysis

The dataset consists of cases distributed all over the world. We split the U.S. based cases into 4 regions, *west*, *midwest*, *northeast* and *south*, according to the United States Census Bureau. We further grouped together all cities outside the U.S. into one single category, namely, *international*. The distribution of cases across different regions is demonstrated in Figure 4. Though we did not explicitly show in the figure, the west is dominated by Arizona and California, due to the large focus of FMG on these two states.

There are several known reasons of disappearance associated with the cases in our dataset including, *accidental*, *bipolar*, *drowning*, *foul play*, *natural*, *runaway*, *self-inflicted*, *staged* and *undetermined*. According to Figure 5, 'foul play' is the dominant reason for disappearance. There are also different number of reporters for each case. The distribution of reporters with respect to the number of cases in which they participated is shown in Figure 6.

Table 1: Description of the dataset

| Name | Value |
|---|---|
| Found Status | |
| Alive | 12 |
| Deceased | 76 |
| Gender | |
| Male | 41 |
| Female | 47 |
| Age | |
| Under 13 | 9 |
| 13 to 30 | 39 |
| 30 and older | 40 |

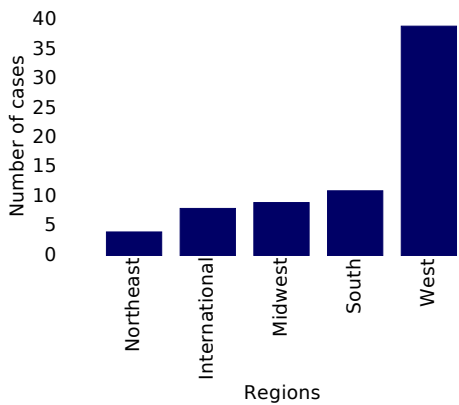

Figure 4: Distribution of the cases across different regions of the US and international.

For the rest of our data analysis, we need to introduce some preliminary notation. We use the random variable $g_A$ to denote if the missing person is found alive or not, so it is either true or false. We shall use $Pr(g_A = \text{True}|o \text{ stated Alive})$ to denote the confidence of the observer $o$ in reporting *Alive*. This confidence value shows the portion of the cases for which $o$ has reported the missing person is *Alive* and he/she was found *Alive*, to the total number of cases for which $o$ has reported *Alive*. Likewise, we compute the confidence of $o$ in reporting *Deceased*. The distribution of the reporters with respect to their confidence values is demonstrated in Figure 7. According to the figure, most reporters' confidence values belong to the ranges of [0.3,0.4) for *alive* and [0.8,0.9) for *deceased* statuses.

We also define the ratio $r_A$ as follows:

$$r_A = \frac{Pr(g_A = \text{True}|\text{observer } o \text{ stated } Alive)}{Pr(g_A = \text{True})} \quad (17)$$

This ratio demonstrates how much the observer $o$ outperformed the prior probability $Pr(g_A = \text{True})$ on *Alive*. Similarly, we use $r_D$ for *Deceased* cases. The distributions of the reporters with respect to $r_A$ and $r_D$ are shown in Figure 8. We note that as most are found dead, it is harder for the reporters to outperform the prior on *Deceased* compared to the *Alive*.

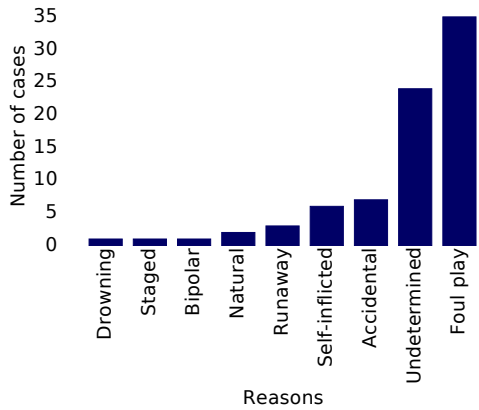

Figure 5: Distribution of the cases with respect to the probable reasons.

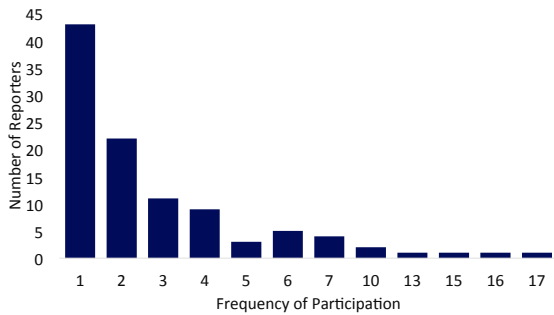

Figure 6: Distribution of frequency of participation.

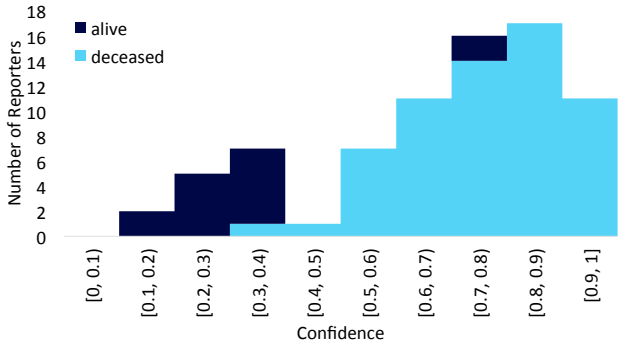

Figure 7: Distribution of all reporters with respect to their confidence values.

## 7. EXPERIMENTAL RESULTS

This section reports on the experiments conducted to validate our approach. We note that the individual cases themselves are not related - hence we are justified in using leave-one-out cross validation in our experiments. Specifically, for each case in the experiments, we learn a *different* model using all of the other cases. We first compare the methods for restricted (without dog) and unrestricted (with dog) searches and then discuss the sensitivity of the parameter.

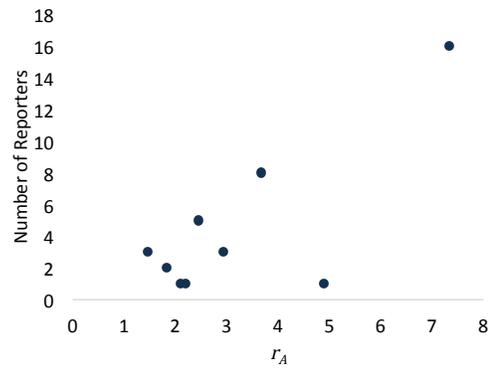
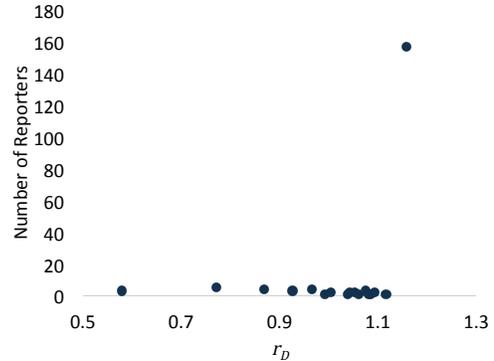

Figure 8: The distributions of the reporters with respect to $r_A$ and $r_D$.

### 7.1 Area Reduction

In this section, we examine how our approach can be used to reduce the area searched by the Find Me Group over the baseline. Figure 9 shows the reduction of area based on our approach (double distance constraint integer program with Algorithm 1 and $\epsilon = 0.1$) when compared to the baseline. We examine this with grid squares of $1 \times 1$ miles and $2 \times 2$ miles. In the 19 cases where the missing person was located, our approach achieved area reduction in 11 cases - reducing the search area by an average by 3 square miles. In the 2 cases where our method caused the search area to increase, the increase was only 1 square mile in each case. This contrasts with the cases where the area was reduced - reducing the search area by up to 9 square miles. For the 11 cases where reduction was experienced, the average reduction was 1.63 miles ($t(19) = 1.25$, $p < 0.11$).

We also examined cases where the size of the grid squares was $2 \times 2$ miles. In the 19 cases, the area reduction achieved by our method was in 14 cases, and by an average by 8.5 square miles. Further, in the 6 cases, our method caused an increase in the search area, however, the increase was 3 square miles on average. Further, for the cases that baseline needs to search areas larger than 20 square miles, our approach reduced the area from 7 to 56. Our method outperformed the baseline in area reduction with an average of 4.21 mile square ($t(20) = 1.19$, $p < 0.13$).

### 7.2 Consideration of Dog Team Detections

The experiments of the previous section illustrated how our approach could reduce the search area over the baseline for standard grid settings. However, in the events that a dog

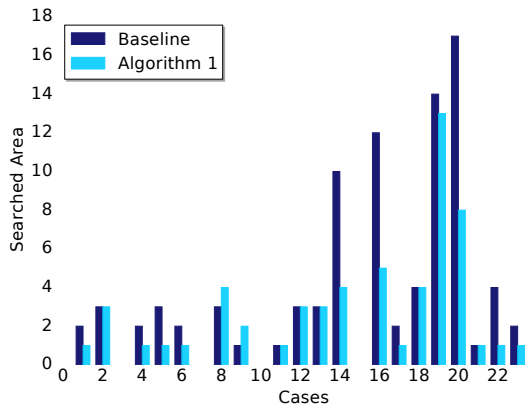

(a) Search area with $1 \times 1$ mile per observation

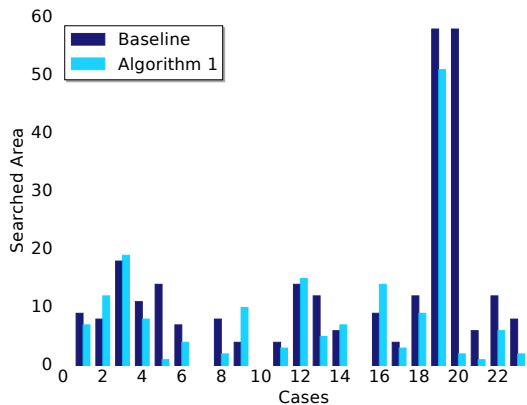

(b) Search area with $2 \times 2$ miles per observation

Figure 9: Searched area until the missing person is located (baseline and Algorithm 1).

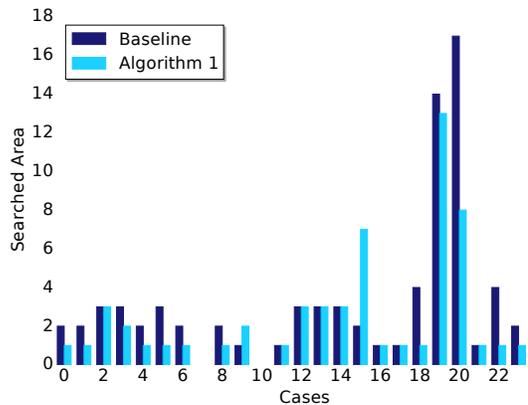

(a) Search area with $1 \times 1$ mile per observation

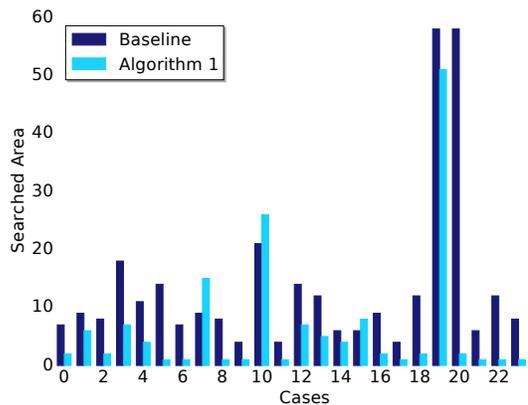

(b) Search area with $2 \times 2$ miles per observation

Figure 10: Searched area with dogs allowed to explore 1 mile beyond the grid (baseline and Algorithm 1).

team detects evidence of the missing person, it may lead to a continued search outside of the assigned grid square. These searches can lead to FMG personnel examining up to a mile outside a designated location. In this section, we consider a grid square settings in the last section, but also allow for an additional mile outside the square to mimic the effect of the dog search team following such a lead. Figure 10 demonstrates the reduction of area based on our approach (double distance constraint integer program with Algorithm 1 and $\epsilon = 0.1$) when compared to the baseline. We investigate the area reduction with grid squares of $1\times1$ miles and $2\times2$ miles. According to Figure 10a, in the 22 cases where the missing person was located, our approach achieved area reduction in 12 cases - reducing the search area by 2 square miles on average. In the 2 cases where our method caused the search area to increase, the increase was only 3 square miles on average. This contrasts with the cases where the area was reduced - reducing the search area by up to 9 square miles. Our method outperformed the baseline in area reduction with an average of 0.86 mile square ($t(22) = 0.8$, $p < 0.22$).

We examined cases where the size of the grid squares was $2\times2$ miles. In the 24 cases, the area reduction achieved by our method was in 21 cases, and on average by 8.85 square miles. In the 3 cases where our method caused the search area to increase, the increase was 4.3 square miles on average. This contrasts with the cases with the reduced search area by up to 56 square miles. Our method outperformed the baseline in area reduction with an average of 7.2 mile square ($t(24) = 1.95$, $p < 0.05$).

### 7.3 Parameter Sensitivity

We compare different values of $\epsilon$ in both double distance constraint integer programs (iterative search resource allocation and non-iterative program). The impact of changing the parameter $\epsilon$ is shown in Figure 11. To do so, we plot the fraction of area searched by our method over the baseline, against the $\epsilon$, for both sizes of $1 \times 1$ and $2 \times 2$. We note that while the extreme values of $\epsilon$ (i.e. 0.0 and 0.5) negatively effected the performance of both approaches, we achieved relatively stable results for intermediate values - noting that the best performance was to set $\epsilon$ equal to 0.1 - which we used in the experiments.

We also studied the performance of our optimization approach without algorithm 1 (i.e. prioritize locations by equation 5 after selecting the values for $\beta_o$ through optimization of 19 with regards to Lines 9-11). The results are depicted in Figure 12. The behavior of the algorithm for different settings of $\epsilon$ were similar to that found with Algorithm 1, the reduction in search area was generally less - and in some

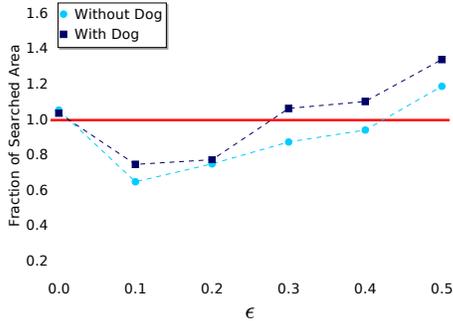

(a) Search area with $1 \times 1$ mile per observation (**Algorithm 1**)

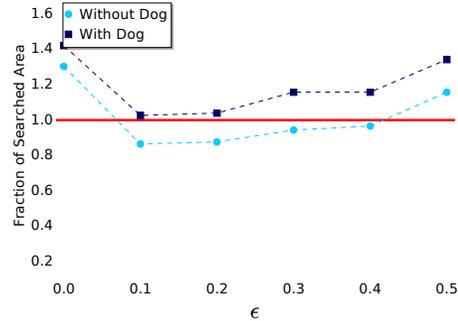

(a) Search area with $1 \times 1$ mile per observation ***not* using Algorithm 1**

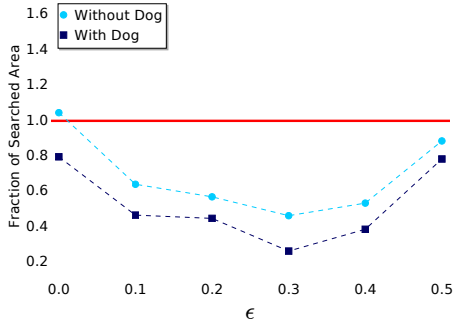

(b) Search area with $2 \times 2$ miles per observation (**Algorithm 1**)

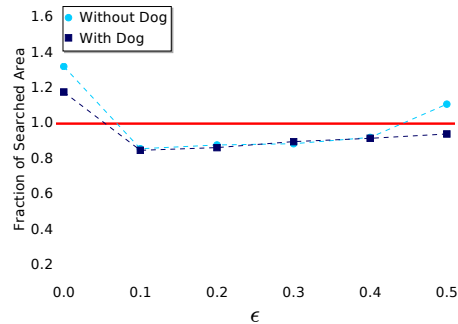

(b) Search area with $2 \times 2$ miles per observation ***not* using Algorithm 1**

Figure 11: Fraction of total area searched across all cases with the iterative search resource allocation approach over the baseline.

Figure 12: Fraction of total area searched across all cases by the double distance constraint integer programming approach (***not* using Algorithm 1**) over the baseline.

cases (i.e. 1x1 mile grid square with use of the dogs) it performed worse.

## 8. RELATED WORK

Recently, there has been some work [19, 20, 21, 23, 9] dealing with geospatial abductive inference introduced in [24]. In [19] for example, authors studied the case of geospatial abduction where there is an explicit adversary who is interested in ensuring that the agent does not detect the partner locations in an attempt to simulating the real-world scenario of insurgents who conduct IED (improvised explosive device) attacks. Another work [20], has adopted geospatial abduction to develop a software tool which applies geospatial abduction to the environment of Afghanistan, to look for insurgent high-value targets, supporting insurgent operations. The work of [21] introduced a variant of the GAPs called region-based GAPs (RGAPs) which deals with the multiple possible definitions of the subregions of the map. Finally, spatial cultural abductive reasoning engine which solves spatial abductive problems was developed in [23]. Aside from introducing GAP, the work of [24] demonstrated the accuracy of proposed framework on real-world dataset of insurgent IED attacks against US forces in Iraq. Further, the work of [9], proposed a technique to reduce the computational cost of point-based GAPs. They presented an exact algorithm for the natural optimization problem of point-based GAPs. Geospatial abduction problems are related to facility location [26] and sensor placement problems [10] in that they identify a set of geo-locations to optimize a cost or reward function. However, there are key differences amongst these various frameworks that arise from the difference between explanation and optimization. See [22] for further discussion on this topic.

Similarly, [1] presents a specific aspect of the well-known qualification problem, namely spatial qualitative reasoning approach, which aims at investigating the possibility of an agent being present at a specific location at a certain time to carry out an action or participate in an event, given its known antecedents. This work is different from both above papers and our study, as it takes on purely logical approach to formalizing spatial qualifications, while our work and other aforementioned studies use geometric and probabilistic techniques. Further, the framework of this paper is tailored specifically for the missing person problem.

Looking beyond geospatial abduction, recent research has demonstrated that GPS (positional) data could be used to learn rich models of human activity [16, 15, 17, 7]. For example, [16, 15, 17], modeled the human interactions and intentions in a fully relational multi-agent setting. They used raw GPS data from a real-world game of capture the flag and Markov logic- a statistical-relational language. Whereas [7] developed a model to simulate the behaviors associated with insurgent attacks, and their relationship with geographic locations and temporal windows.

At first glance, one may think our work is similar to [10], in that they identify a set of geo-locations to optimize a cost or reward function. However, as described, there are key differences amongst these various frameworks that arise from the difference between explanation and optimization.

## 9. CONCLUSION

In this paper, we have introduced the Missing Person Intelligence Synthesis Toolkit (MIST) which leverages a data-driven variant of geospatial abductive inference. MIST can rank-order the set of search locations provided by a group of experts. The experimental results showed that our approach is able to reduce the total search area by a total of 31 square miles for standard searched and by 19 square miles when dog team assets obtain a detection. This reduction will make FMG locating missing persons faster while saving in direct and indirect cost. At the time of this writing, we have initiated support to FMG with MIST for an active case. FMG will use MIST's ranking of search locations for this ongoing operation.

Our future plans include utilizing a probabilistic variant of the feasibility function, applying other features such as missing person's region, age, gender to the model and extending our toolkit to be able to solve other problems such as human trafficking.

### Acknowledgement

This work was funded by the Find Me Group.